\documentclass[english]{article}
\usepackage{eucal}
\usepackage[utf8]{inputenc}
\usepackage[T1]{fontenc}
\usepackage{babel}
\usepackage{amsmath}
\usepackage{amsfonts}
\usepackage{csquotes}
\usepackage{amstext}
\usepackage{amssymb}
\usepackage{graphicx}
\usepackage{comment}
\usepackage{subcaption}
\usepackage{fancyhdr}
\usepackage{siunitx}
\usepackage{hyperref}
\hypersetup{
    colorlinks=true,
    linkcolor=blue,
    filecolor=magenta,      
    urlcolor=cyan,
}
\usepackage{graphicx}

\usepackage{biblatex}
\begin{document}

\title{Deep reinforcement learning-based image classification achieves perfect testing set accuracy for MRI brain tumors with a training set of only 30 images}

\author{ \textbf{Joseph Stember}$^1$
\and 
\textbf{Hrithwik Shalu}$^2$}
\maketitle
\thispagestyle{fancy}
\noindent
\textsuperscript{1}Memorial Sloan Kettering Cancer Center, New York, NY, US, 10065 
\\
\textsuperscript{2}Indian Institute of Technology, Madras, Chennai, India, 600036
\\
\noindent
\textsuperscript{1}joestember@gmail.com
\\
\textsuperscript{2}lucasprimesaiyan@gmail.com 
\\

\begin{abstract}

\indent \textit{Purpose} Image classification may be the fundamental task in imaging artificial intelligence. We have recently shown that reinforcement learning can achieve high accuracy for lesion localization and segmentation even with minuscule training sets. Here, we introduce reinforcement learning for image classification. In particular, we apply the approach to normal vs. tumor-containing 2D MRI brain images. 

\indent \textit{Materials and Methods}  We applied multi-step image classification to allow for combined Deep Q learning and TD(0) Q learning. We trained on a set of $30$ images ($15$ normal and $15$ tumor-containing). We tested on a separate set of $30$ images ($15$ normal and $15$ tumor-containing). For comparison, we also trained and tested a supervised deep learning classification network on the same set of training and testing images. 

\indent \textit{Results}
Whereas the supervised approach quickly overfit the training data and as expected performed poorly on the testing set ($57 \%$ accuracy, just over random guessing), the reinforcement learning approach achieved an accuracy of  $100 \%$.

\indent \textit{Conclusion}
We have shown a proof-of-principle application of reinforcement learning to classification of brain tumors. We achieved perfect testing set accuracy with a training set of merely $30$ images.

\end{abstract}

\pagebreak

\section*{Introduction}

Image classification may be the fundamental task of artificial intelligence (AI) in radiology \cite{mcbee2018deep,saba2019present,mazurowski2019deep,weikert2020practical}. Essentially all AI classification currently practiced, like the tasks of localization and segmentation, falls within the category of supervised deep learning. 

Supervised deep learning (SDL) classification research necessitates acquiring and often pre-processing a large number of appropriate images consisting of the various categories of interest. Typically, hundreds and often thousands or tens of thousands of images are needed for successful training. Radiologists must label each image, specifying the class/category to which each belongs. Often to increase (somewhat artificially) the training set, augmentation operations are performed. Once enough data is gathered, processed, and labeled, it can be fed into a convolutional neural network (CNN) that predicts image classes from the output.

SDL in general suffers from three crucial limitations that we have sought to address through reinforcement learning:

\begin{enumerate}
  \item As above, SDL requires many curated and labeled images to train effectively. 
  \item Lack of generalizability renders SDL susceptible to fail when applied to images from new scanners, institutions, and/or patient populations \cite{wang2020inconsistent,goodfellow2014explaining}. Importantly, this limits clinical utility.
  \item The "black box" phenomenon of non-understandable AI, in which algorithm opaqueness hinders trust of the technology. Trust is paramount for consequential health care decisions. Obscurity also limits contributions from those without extensive AI experience but with advanced domain knowledge (e.g., radiologists or pathologists) \cite{buhrmester2019analysis,liu2019comparison}.
\end{enumerate}

In recent work \cite{stember2020deep,stember2020reinforcement,stember2020unsupervised}, we have introduced the concept of radiological reinforcement learning (RL). We showed that RL can address at least two of the above challenges when applied to lesion localization and segmentation. 

Another fundamental task of deep learning is classification. Classifying an image into two generalized categories (normal and abnormal) is widely viewed as a fundamental task \cite{tang2018canadian}. As such, and as two-outcome classification can be generalized to any number of different classes, we sought here to use RL for two-category classification as a proof-of-principle.

\section*{Methods}

\subsection*{Data collection}

We collected $60$ two-dimensional image slices from the BraTS 2020 Challenge brain tumor database \cite{menze2014multimodal,bakas2017advancing,bakas2018identifying}. As we used publicly available images, IRB approval was deemed unnecessary for this study. All images were T1-weight post-contrast and obtained at the level of the lateral ventricles. Of the $60$ image slices, $30$ were judged by a neuroradiologist (JNS, $2.5$ years of clinical experience) to be normal in appearance. The other $30$ contained enhancing high grade gliomas. We employed $30$ images ($15$ normal and $15$ tumor-containing) for the training set. The other $30$ ($15$ normal and $15$ tumor-containing) were assigned to the testing set.

\subsection*{Reinforcement learning environment, definition of states, actions, and rewards}

\begin{figure}
\centering
\includegraphics[width=11cm,height=7.5cm]{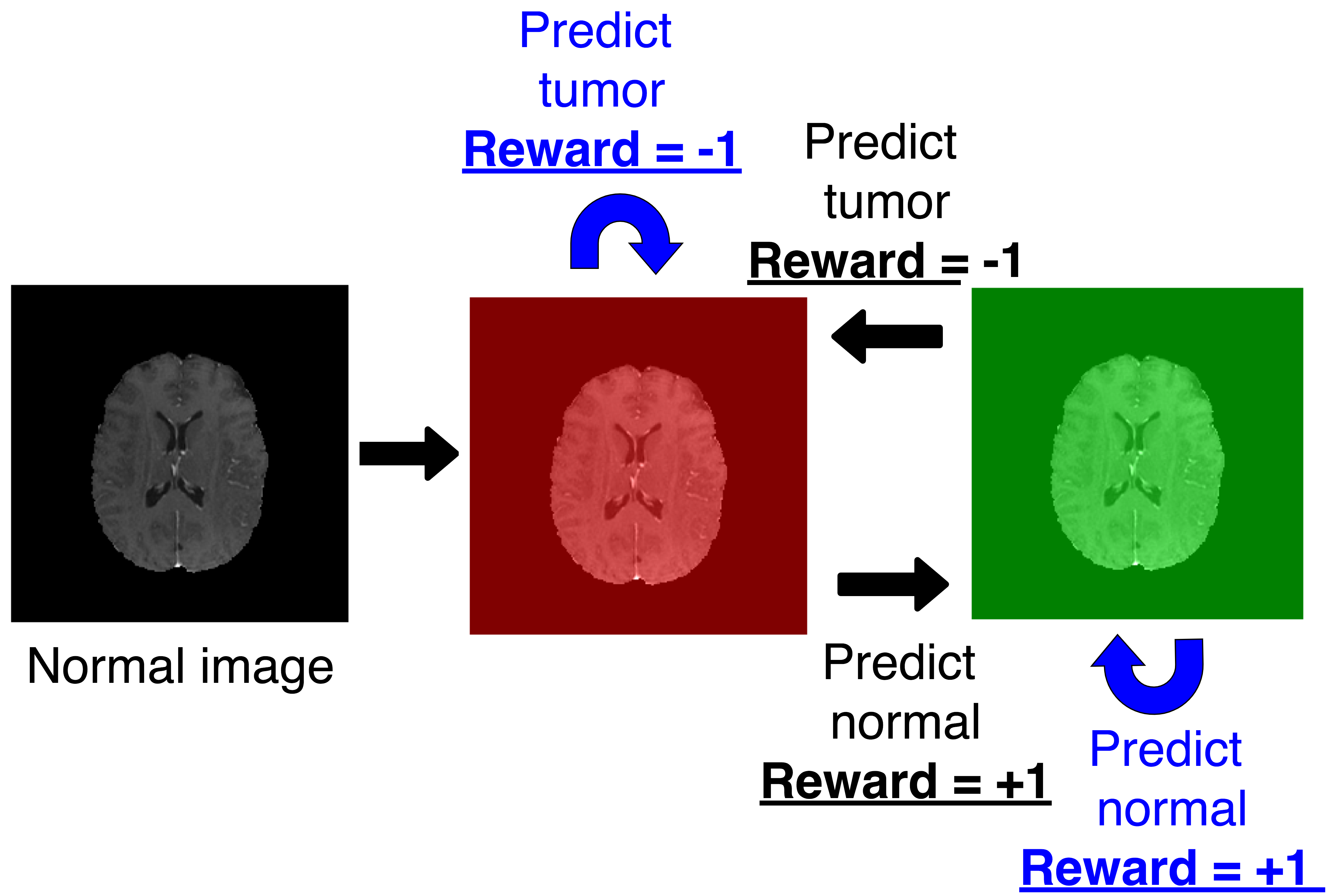}
\caption{ Markov decision process for a normal image. }
\label{fig:MDP_normal}
\end{figure}

\begin{figure}
\centering
\includegraphics[width=11cm,height=7.5cm]{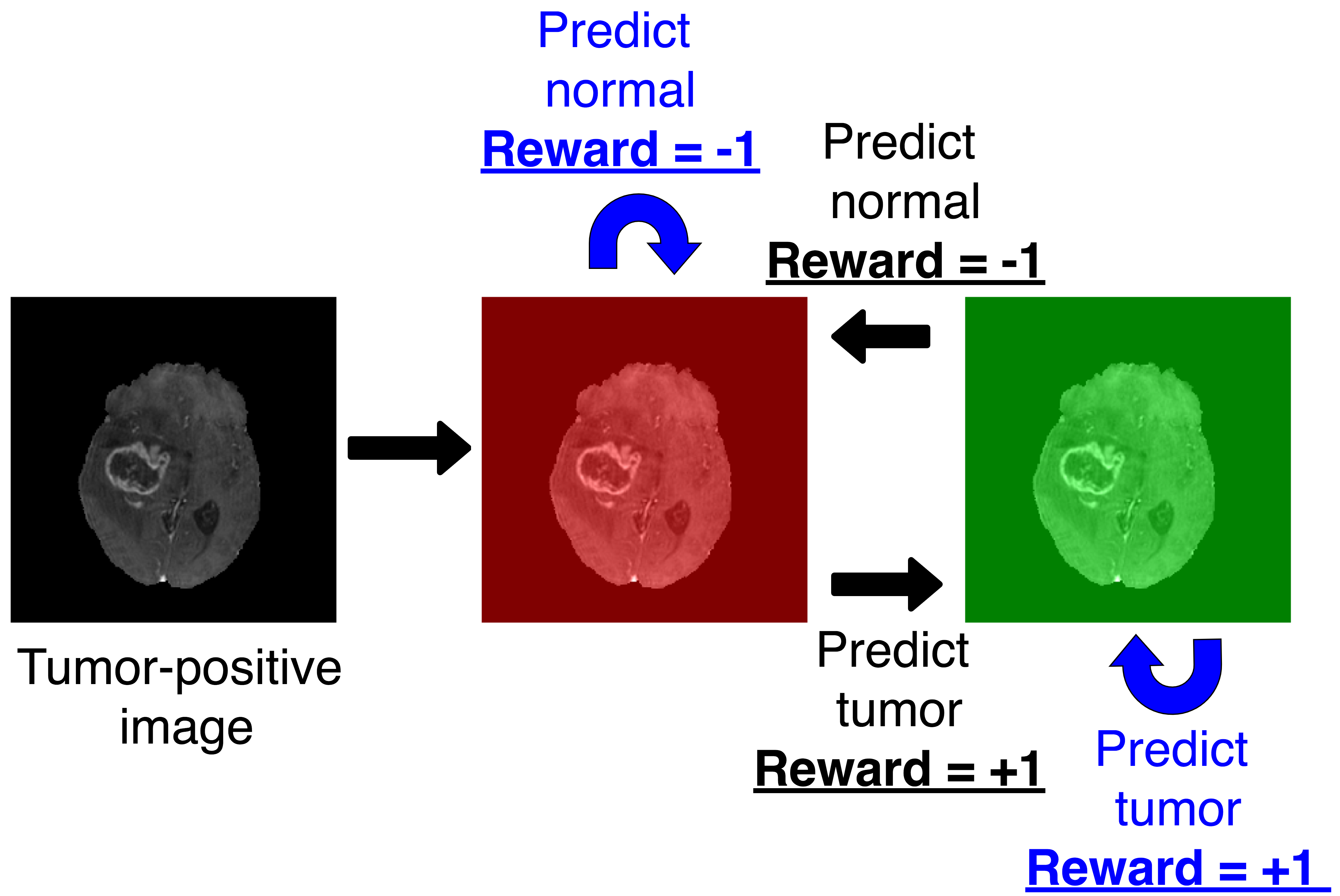}
\caption{ Markov decision process for a tumor-containing image. }
\label{fig:MDP_tumor}
\end{figure}

In keeping with standard discrete-action RL, the framework we used to produce optimal policy is the Markov Decision Process (MDP). The MDP for our system is illustrated in Figures \ref{fig:MDP_normal} and \ref{fig:MDP_tumor}. The former illustrates the MDP for a normal image. The grayscale representation of the image is overlaid in red or green to represent the states. For the purpose of illustration, in these figures, the alpha value for transparency was set higher than in our actual calculations (0.5 vs. 0.1) to make the colors appear sharper. 

The grayscale image is converted to red-overlay, the latter representing initial state, $s_1$. At each step in the episode ($5$ steps per episode during training), the agent takes an action. That action, represented by either $0$ or $1$, predicts whether the image belongs to the normal or tumor-containing class, respectively. If the action predicts the correct class, the next state is green-overlay. If it predicts the wrong class, which in the case of a normal image would be to predict tumor-containing, the state would remain red-overlay or flip from green-overlay to red-overlay. The reverse is true for the tumor-containing image, shown in Figure \ref{fig:MDP_tumor}. If the action is $0$, predicting normal image, the next state is red-overlay, whereas if it makes the correct prediction of tumor-containing (action $a_t = 1$), the next state $s_{t+1}$ is green-overlay. 

The agent is provided a reward of $+1$ for taking the correct action / class prediction, and is penalized with a reward of $-1$ for a wrong action / prediction. This is also shown in Figures \ref{fig:MDP_normal} and \ref{fig:MDP_tumor}. The ultimate goal of RL training is to maximize total cumulative reward. 

\subsection*{Training}

As in our prior work \cite{stember2020deep,stember2020reinforcement,stember2020unsupervised}, we employed an off-policy $ \epsilon $-greedy strategy that permits exploration of non-optimal states. This allows the agent to learn about the environment by exploring states with a randomness that over time gives way to more deterministic, on-policy actions as the agent learns the environment and gets closer to an optimal policy. We again used $ \epsilon = 0.7 $ for initial random sampling, slowly decreasing to the the minimal value $ \epsilon_{\text{min}} = 1 \times 10^{-4} $. 

\begin{figure}
\centering
\includegraphics[width=11.5cm,height=8cm]{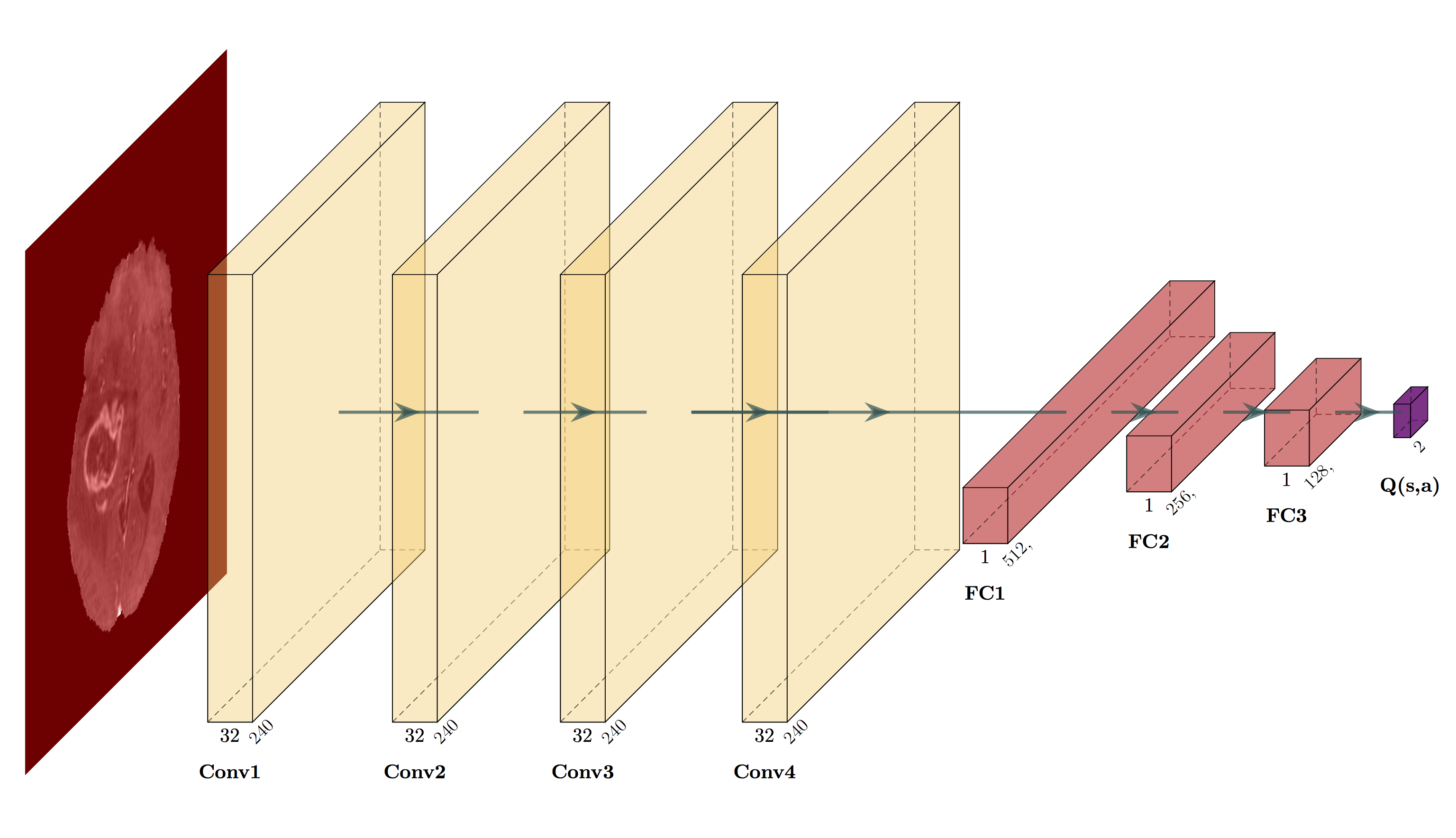}
\caption{ Deep Q network (DQN) architecture. }
\label{fig:DQN_architecture}
\end{figure}

We followed the protocol from our prior work by also employing a Deep $Q$ network (DQN) in tandem with $TD(0)$ $Q$-learning. The former computes actions from input state via the DQN, which is basically a CNN, displayed in Figure \ref{fig:DQN_architecture}. The architecture is essentially identical to that used in our recent work for lesion localization and segmentation. The two outputs from this network are the state-action value function $Q(s_t,a_t)$. $Q(s_t,a_t)$ computes the value of taking action $a_t$ in state $s_t$. 

Our agent samples states and learns about the environment via the reward, which refines the version of $Q$ that we call $Q_{\text{target}}$. As in our recent work, we sampled via temporal difference $Q$ learning in its simplest form: $TD(0)$. Doing so, for time $t$, we updated $Q_{\text{target}}^{(t)}$ via the Bellman Equation:
\begin{equation} 
    Q_{\text{target}}^{(t)} = r_t + \gamma max_a Q(s_{t+1},a) \text{,}
\label{bellman_eqn}
\end{equation}
where $\gamma = 0.99$ is the discount factor, which reflects the relative importance of immediate vs. most distant future rewards, and $max_a Q(s_{t+1},a)$ is equivalent to the state value function $V(s_{t+1})$. 

With repeated sampling by Equation \ref{bellman_eqn}, $Q_{\text{target}}^{(t)}$ eventually converges toward the optimal $Q$ function, $Q^{\star}$. As implied by the name, $Q_{\text{target}}$ serves as the target in the DQN from Figure \ref{fig:DQN_architecture}. Hence, minimizing the loss between the network output $Q_{\text{DQN}}$ and $Q_{\text{target}}$ via backpropagation in combination with sampling the environment through the Bellman Equation, we arrive at 
\begin{equation}
    \displaystyle{\lim_{t \to \infty}} \left( Q_{\text{DQN}}(t) \right) = Q^{\star}.
\end{equation}
By following the above-described process, we arrived at a CNN/DQN approximation of the optimal $Q$ function. This allowed us to act as per the optimal policy thereafter. We can do so in state $s$ by selecting action $a = max_a Q(s,a)$, where $Q(s,a)$ is produced by a forward pass of the trained DQN on input state $s$.

As described in earlier work, the data on which the DQN trains is obtained by the "memory" of prior state-action-next state-reward tuples, stored in a so-called transition matrix $\mathbb{T}$. $\mathbb{T}$ is of size $4 \times N_{\text{memory}}$, $N_{\text{memory}}$ being the replay memory buffer size. We used the value of $N_{\text{memory}} = 1,500$ based on recent experience, noting that this tends to produce enough samples to represent the agent's experience, while not overwhelming the CPU capacity. During DQN training, batches of size $n_{\text{batch}} = 32$ transitions are randomly sampled from $\mathbb{T}$.  

At each step of training, a normal or tumor-containing image is sampled randomly with equal probability. Each episode of training consists of five steps as per Figures \ref{fig:MDP_normal}, \ref{fig:MDP_tumor}. We trained for a total of $300$ episodes. Regarding the DQN, we used the Adam optimizer with learning rate of $1 \times 10^-4$ and using mean squared error loss between $Q_{\text{DQN}}$ and $Q_{\text{target}}$. We used $3 \times 3$ convolutional kernels, with weights initialized by the standard Glorot initialization. 

At each step of each episode, a new row of $\mathbb{T}$ is calculated, a new value of $Q_{\text{target}}$ can be computed and compared to $Q_{\text{DQN}}$ for an additional element of the loss, and another iteration of forward and backpropagation can occur. Hence the DQN is trained for a total of $N_{\text{episode}} \times n_{\text{steps}} = 300 \times 5 = 1,500$ "epochs." In order to keep $N_{\text{memory}}$ fixed at $1,500$, older transitions are discarded from $\mathbb{T}$ in favor of newly computed transitions.

\subsection*{Supervised deep learning (SDL) classification for comparison}

To compare SDL and RL-based classification, we trained the same $30$ training set images with a CNN with architecture essentially identical to that of the DQN. The CNN also consisted of convolutional layers followed by elu activation employing $3 \times 3$ filters. As for the DQN, this was followed by three fully connected layers. The network outputs a single node, which is passed through a sigmoid activation function given the direct binary nature of the CNN's prediction (normal vs. tumor-containing). The loss used here is binary cross entropy. Other training hyperparameters were the same as for the DQN. The supervised CNN was trained for $300$ epochs. 

\section*{Results}

\begin{figure}
\centering
\includegraphics[width=11.5cm,height=8.5cm]{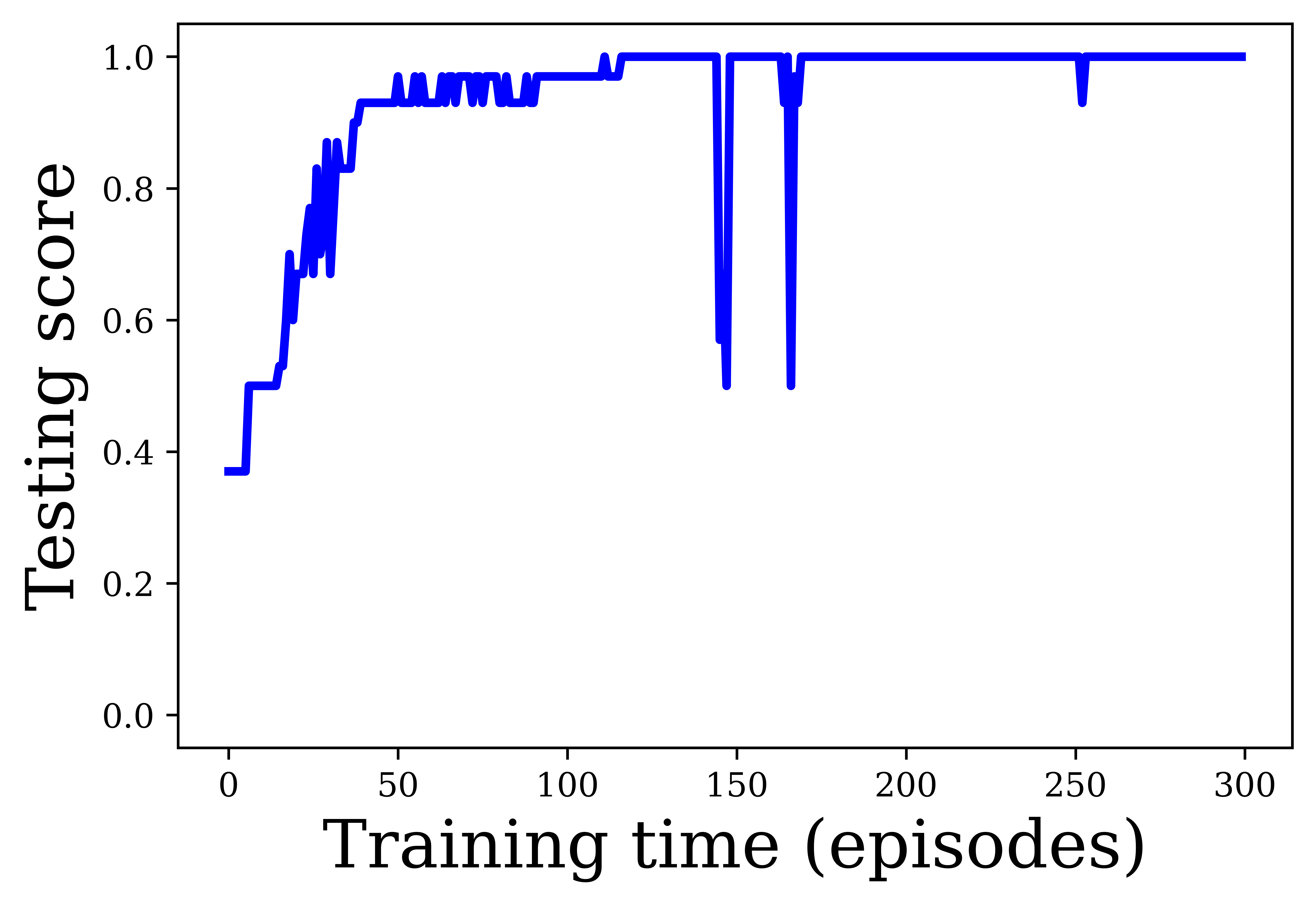}
\caption{ Testing set accuracy during the course of reinforcement learning training. }
\label{fig:test_acc}
\end{figure}

The testing process for the RL approach consisted of making a single step prediction on the testing set images with initial states of red-overlay. Figure \ref{fig:test_acc} shows the testing set accuracy as a function of training time. We can see steady learning that generalizes to the testing set, essentially plateauing at $100 \%$ accuracy within $200$ episodes. No strict analogue for training time exists between RL and SDL. However, we employ the most analogous measures of episodes and epochs, respectively.

\begin{figure}
\centering
\includegraphics[width=11.5cm,height=8.5cm]{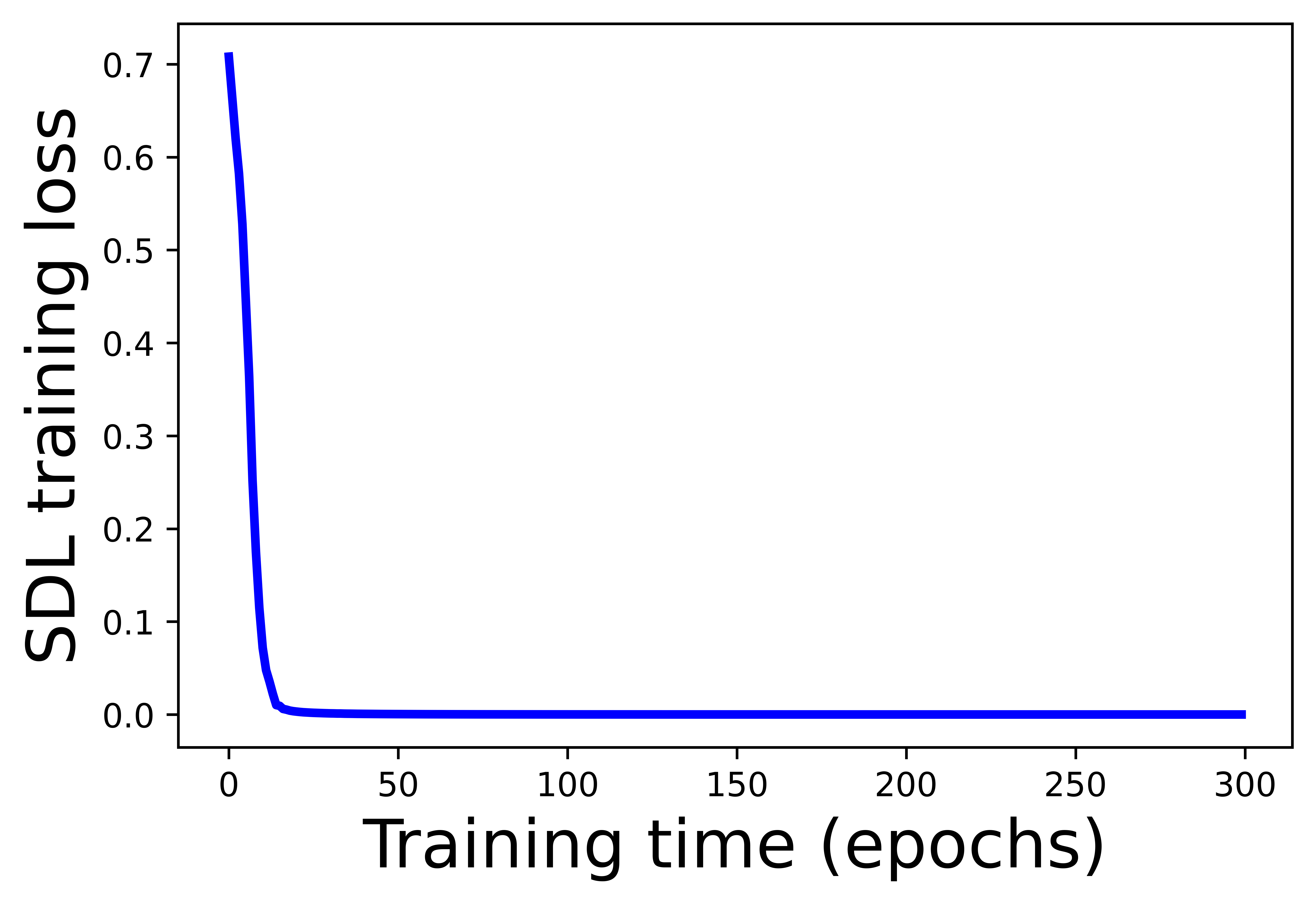}
\caption{ Unsupervised learning training set loss. The method overfits the small training set as it successively decreases loss. }
\label{fig:train_loss}
\end{figure}

The loss during training of the supervised CNN is shown in Figure \ref{fig:train_loss}. The network is seen to train properly, with an initial sharp drop in loss as it quickly overfits the exceedingly small training set.

\begin{figure}
\centering
\includegraphics[width=11.5cm,height=8.5cm]{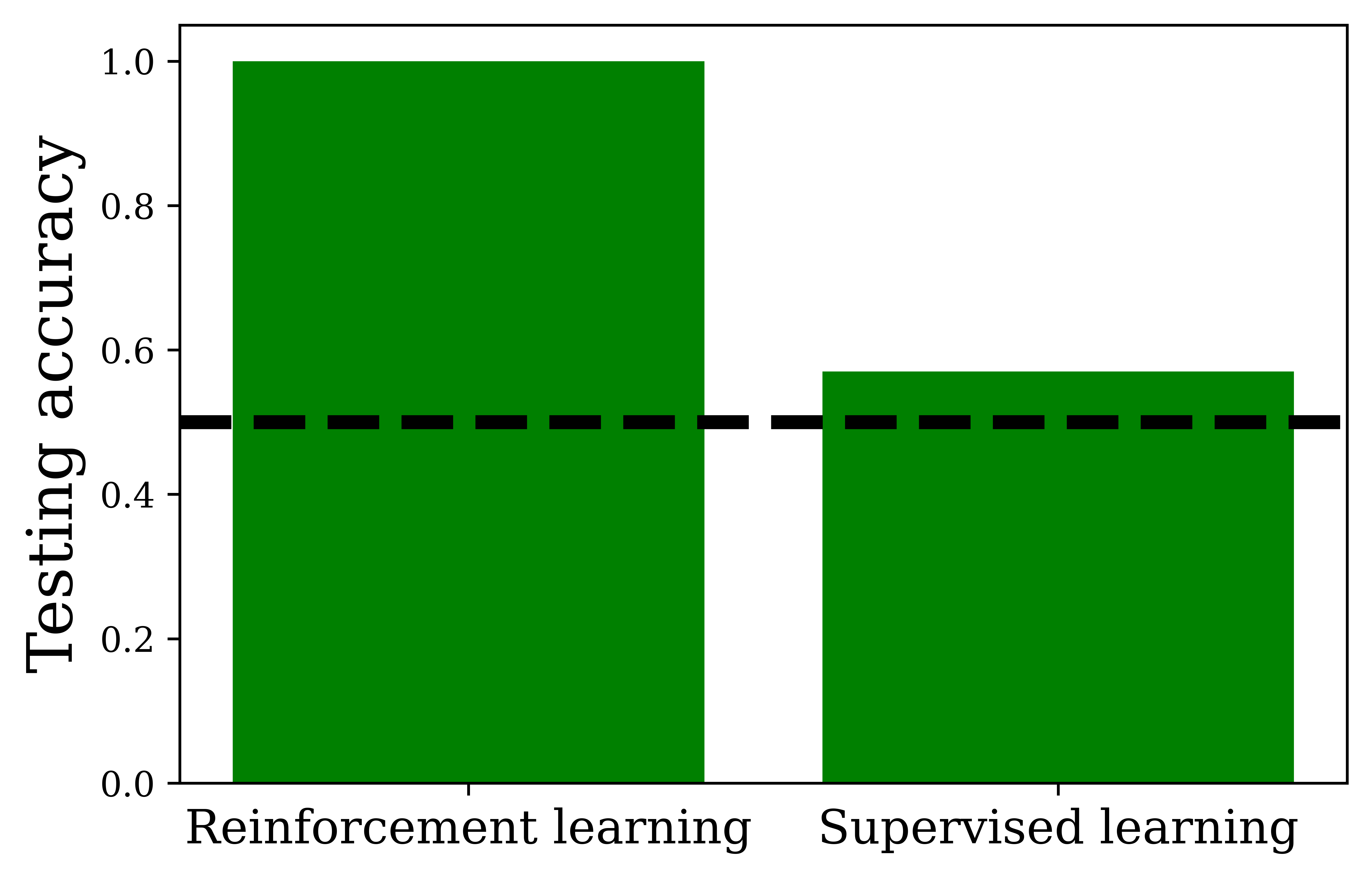}
\caption{ Bar plot showing the testing set accuracy (bar height) for reinforcement learning vs. supervised deep learning. The dashed horizontal line corresponds to $50 \%$ accuracy, equivalent to random guess.  }
\label{fig:acc_comp}
\end{figure}

We compare the testing set accuracy of RL and SDL in Figure \ref{fig:acc_comp}. In contrast to the $100 \%$ accuracy of RL, SDL has a mere $57 \%$ accuracy for the testing set, just above a $50 \%$ random guess. This is due to the fact that SDL is bound to overfit the very small training set whereas RL learns general principles that can be applied with success to the separate testing set.

\section*{Discussion}

We have shown that, when applied to a small training set, reinforcement learning vastly outperforms the more "traditional" supervised deep learning for lesion localization \cite{stember2020deep,stember2020reinforcement}, segmentation \cite{stember2020unsupervised}, and  classification. 

Our use of two-dimensional images is one limitation of the study. Furthermore, we have only used two classes. Future work will extend the approach to full three-dimensional images from our institution and will generalize to multiple image classes / classifications.

\section*{Conflicts of interest}

The authors have pursued a provisional patent based on the work described here.

\printbibliography

\end{document}